%% file: ms.tex
\documentclass[10pt,twocolumn,letterpaper]{article}

\usepackage{iccv}
\usepackage[dvipsnames]{xcolor}
\usepackage{times}
\usepackage{epsfig}
\usepackage{graphicx}
\usepackage{amsmath}
\usepackage{amssymb}
\usepackage{svg}
\usepackage{calc}
\usepackage{flushend}

\usepackage{subfig}
\include{macro}

\usepackage[pagebackref=true,breaklinks=true,letterpaper=true,colorlinks,bookmarks=false]{hyperref}

\iccvfinalcopy 

\def\iccvPaperID{****} 

\ificcvfinal\fi

\begin{document}

\title{Content-Consistent Generation of Realistic Eyes with Style}


\author{
Marcel B\"uhler\textsuperscript{1},
Seonwook Park\textsuperscript{1},
Shalini De Mello\textsuperscript{2},
Xucong Zhang\textsuperscript{1},
Otmar Hilliges\textsuperscript{1} \\
\textsuperscript{1}ETH Z{\"u}rich,\qquad\textsuperscript{2}NVIDIA\\
{\tt\small \{buehlmar,spark,\,zhangxuc,\,otmarh\}@ethz.ch;\quad shalinig@nvidia.com}
}


\maketitle
\ificcvfinal\thispagestyle{empty}\fi

\input{01_abstract}
\input{02_introduction}

\input{03_related_work}

\input{04_method}
\input{06_results}

\input{07_conclusion}

{\small
\bibliographystyle{ieee_fullname}
\bibliography{references}  

}

\end{document}

%% file: macro.tex
\usepackage{xspace}
\usepackage{multirow}
\usepackage{ifthen}
\usepackage{algorithmicx}
\usepackage{algorithm}
\usepackage{algpseudocode}
\usepackage{colortbl}  
\usepackage{graphicx}
\usepackage{changes}

\usepackage{paralist}
\usepackage{dblfloatfix}

\graphicspath{{figures/}}

\newcommand{\bv}[1]{\mathit{\mathbf{#1}}}

\newcommand{\ganmethod}[0]{Seg2Eye\xspace}

\makeatletter
\newcommand\semitiny{\@setfontsize\notsotiny{6.31415}{7.1828}}
\makeatother

\long\def\ignorethis#1{}

\definecolor{lightgray}{rgb}{0.92,0.92,0.92}
\definecolor{gray}{rgb}{0.35,0.35,0.35}
\definecolor{darkgreen}{rgb}{0.5,0.5,0}
\definecolor{MyBlue}{rgb}{0,0.2,0.8}
\definecolor{MyRed}{rgb}{0.8,0.2,0}
\definecolor{MyGreen}{rgb}{0.0,0.5,0.1}
\definecolor{MyGray}{rgb}{0.4,0.4,0.4}
\definecolor{airforceblue}{rgb}{0.36, 0.54, 0.66}

\newlength\paramargin
\newlength\figmarginstart
\newlength\figmargin
\newlength\subfigmargin
\newlength\secmargin
\newlength\subsecmargin
\newlength\tabmargin
\newlength\tabmarginstart
\newlength\eqmargin
\newlength\algmargin

\usepackage{array}
\newcolumntype{L}[1]{>{\raggedright\let\newline\\\arraybackslash\hspace{0pt}}m{#1}}
\newcolumntype{C}[1]{>{\centering\let\newline\\\arraybackslash\hspace{0pt}}m{#1}}
\newcolumntype{R}[1]{>{\raggedleft\let\newline\\\arraybackslash\hspace{0pt}}m{#1}}



%% file: 01_abstract.tex
\begin{abstract}
Accurately labeled real-world training data can be scarce, and hence recent works adapt, modify or generate images to boost target datasets.
%
However, retaining relevant details from input data in the generated images is challenging and failure could be critical to the performance on the final task.
In this work, we synthesize person-specific eye images that satisfy a given semantic segmentation mask (content), while following the style of a specified person from only a few reference images.
We introduce two approaches, (a) one used to win the OpenEDS Synthetic Eye Generation Challenge at ICCV 2019, and (b) a principled approach to solving the problem involving simultaneous injection of style and content information at multiple scales.
Our implementation is available at \url{https://github.com/mcbuehler/Seg2Eye}.

\end{abstract}

%% file: 02_introduction.tex
\section{Introduction}
\begin{figure}[tbp]
  \centering    
  \vskip -1mm
\includegraphics[width=\columnwidth]{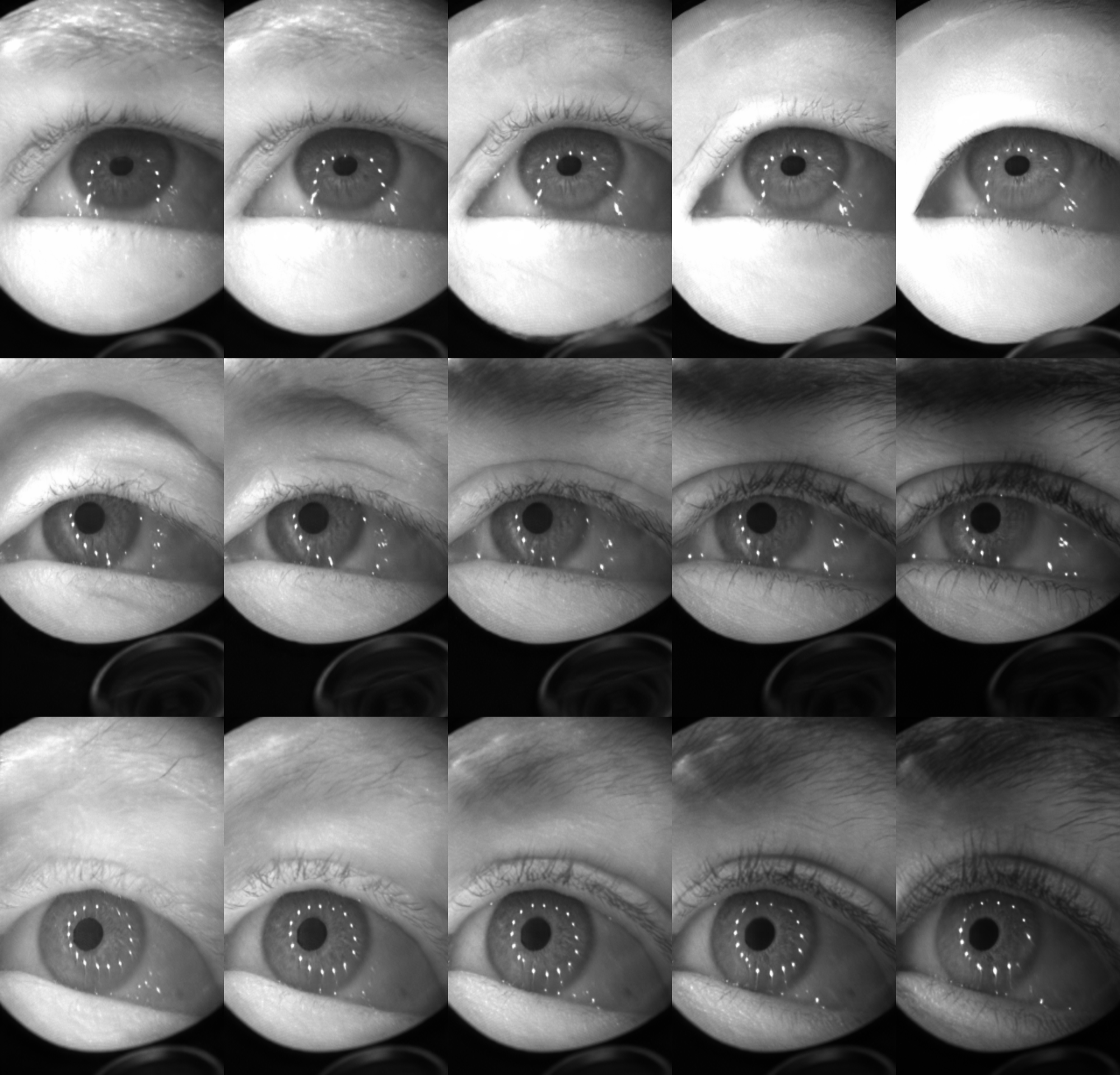}
    \vskip -2mm
    \caption{Walking the style latent space in our proposed method, \ganmethod. We extract latent style codes from two people and show the decodings of their linear interpolation.}
    \label{fig:interpolation}
    \vskip -2mm
\end{figure}

Recent generative models are capable of synthesizing realistic images by using adversarial methods.
However, realism is not the only requirement for computer vision research, as content and style can play important roles for specific tasks, such as regression tasks which require high accuracy, e.g. hand joints regression and eye gaze estimation.
In this paper, we study the task of generating realistic near-eye images while preserving the content defined by a semantic segmentation mask, and style defined by a few images from a target person.
We propose two methods to tackle this task.
Our first method uses image refinement and is the winning solution of the OpenEDS Synthetic Eye Generation Challenge 2019\footnote{\scriptsize\url{https://research.fb.com/programs/openeds-challenge}},
and our second method is a novel architecture for ensuring preservation of desired content and style.
%
%
However, due to optimizing for a very specific error metric, the generated images show blurry regions. Therefore, we propose another more principled method for image synthesis that produces realistic high-quality images that still satisfy both content and style, and furthermore allows for an interpolation between styles.
This method, \ganmethod (Style and Semantic Segmentation preserving GAN), uses content-preserving spatially adaptive normalization blocks (SPADE) \cite{Park2019SPADE} alongside style-preserving adaptive instance normalization layers (AdaIN) \cite{Huang2017AdaIN,Huang2018ECCV,karras2018stylegan}, to inject both content and style information at different feature map scales. 
It is simple yet highly effective when applied in conjunction with a style consistency loss. 
In addition, style injection in \ganmethod is performed from latent embeddings of multiple reference style images from the target person. 
This allows for control as well as a sampling from the learned latent space for synthesizing entirely new people (see Figure~\ref{fig:interpolation}).

%% file: 03_related_work.tex
\section{Related Work}
%

\paragraph{Gaze Estimation.} Recent works in gaze estimation modify existing synthetic or real eye images through domain randomization \cite{Park2018ETRA}, style transfer \cite{Shrivastava2017CVPR,Sela2017arXiv,Lee2018ICLR} and gaze re-direction \cite{Ganin2016ECCV,wood2018gazedirector,Yu2019CVPR,He2019ICCV,Park2019ICCV} to yield data for training more robust and accurate models.
However, the style transfer methods can be poor in preserving content (i.e. eye shape), and re-direction methods can struggle with extrapolating from gaze directions available during training.

\paragraph{Image Translation.} Prior art in image translation often train domain-specific models for cross-domain style transfer \cite{isola2017pix2pix,wang2018pix2pixhd}, use  
Adaptive Instance Normalization (AdaIN) to allow real-time control of visual style \cite{Li2017AAAI,Huang2017AdaIN}, and more recently inject style \cite{karras2018stylegan} or content information (via SPADE blocks \cite{Park2019SPADE}) at multiple feature map scales.
Our proposed \ganmethod approach combines the power of AdaIN and SPADE blocks for control of the style and content of generated eye images respectively.

\section{Dataset}
The provided dataset for the challenge, OpenEDS \cite{Garbin2019arXiv}, is a collection of near infra-red eye images from $152$ people captured by a virtual-reality headset (similar to \cite{Kim2019CHI}) and includes segmentation map annotations for pupil, iris and sclera regions for a subset of $12,759$ images and corneal topography data for $143$ of the $152$ participants.
The unlabeled dataset has two subsets: a ``generative'' subset with $252,690$ images, and a ``sequence'' dataset with $91,200$ frames (from $1.5$ seconds videos collected at $200$Hz).
%

%% file: 04_method.tex
\section{Method}
\begin{figure*}[htbp]
    \vskip -2mm
      \includegraphics[width=\textwidth]{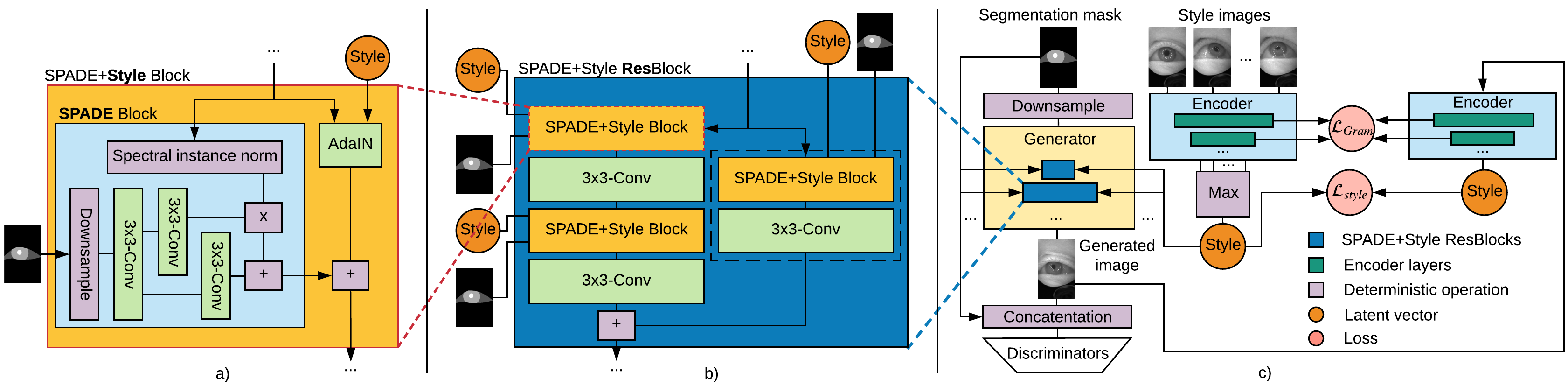}
    \vskip -3mm
    \caption{\ganmethod architecture. (a) describes our novel SPADE+Style Block, which combines Adaptive Instance Normalization (AdaIN) and Spatially Adaptive Normalization (SPADE) to allow for simultaneous style and content injection into the generator at multiple scales. (b) shows how the SPADE+Style Block is used in our ResBlocks, and (c) shows the overall architecture. The generator takes a downsampled segmentation mask and a latent style code as input and produces a synthetic image. The style code is calculated from several style images. We illustrate losses for style consistency, but omit the discriminator losses for brevity. 
    %
    \label{fig:spadestyleall}
    }
    \vskip -4mm
\end{figure*}



In this section, we describe how we won the OpenEDS challenge, and our suggested approach to generating realistic eye images while preserving both content and style. 

\subsection{Eye Segmentation and Similarity Ranking}
\label{sec:methodnn}
The OpenEDS dataset exhibits two modes, which we assume to come from left vs. right eyes. Feeding a left image in order to generate a synthetic image of the right eye impacts performance due to sources of high error (glints, black regions from the head mount) not being apparent in segmentation maps nor consistently in the unlabeled images. Hence, there is a need to carefully select images that come from the same mode. In addition, the more similar the selected unlabeled image is to the target image, the easier it is for the network to produce high-quality output.

In order to find unlabeled images that are similar to a segmentation mask, we train a DeepLab v3+ network \cite{Chen2016DeepLab:CRFs,chen2017rethinking} to predict pseudo-labels (segmentation masks) on the unlabeled dataset and compare them to the target segmentation mask.  We use mean squared error as similarity measure.
%
The unlabeled images are then ranked by the similarity of their predicted segmentation mask with the target segmentation mask. 
We found the matching to perform better when coloring the segmentation masks by the mean value of the respective region across all persons in the training set. 

The output of this step is a ranked list of unlabeled images of the same person for each segmentation mask. This ranking is later used to either modify a similar image to satisfy a test segmentation mask (Section \ref{sec:refiner}), or generate a new image (Section \ref{sec:spadestyle}).


\subsection{Refiner Network}
\label{sec:refiner}
%


\paragraph{Learning a Residual Map.}
The refiner $\mathcal{R}_\Theta$ is a DeepLab v3+ network \cite{chen2017rethinking} with modified inputs and a different loss function. It learns a residual map from the target segmentation map $M_T$, a similar reference image $I$ from the same mode and its pseudo-label $M_I$. The predicted residual map is added to the reference image $I$ in order to produce the final output image $\hat I$. We train the refiner end-to-end with mean L2 error as the optimization objective.
\begin{equation}
    \hat T  = I + \mathcal{R}_\Theta (M_T, M_I, I)
\end{equation}


%
%

\subsection{\ganmethod Network}
\label{sec:spadestyle}
While the refinement approach (Section \ref{sec:refiner}) wins the challenge 
the images produced are not comparable in visual fidelity to the near IR eye images in the dataset (Figure \ref{fig:resultsrefiner}).
Furthermore, the previous approach is unable to synthesize new styles (or person identities). Hence, we propose a generative adversarial network approach, which learns latent style embeddings of unlabeled images and merges them in order to produce photo-realistic outputs.

The basis of our method is a mixture between GauGAN \cite{Park2019SPADE} and StyleGAN \cite{karras2018stylegan}. GauGAN is a recently proposed generative adversarial network (GAN) with strong segmentation mask (content) consistency, whereas StyleGAN learns a suitable latent representation of style and injects it via AdaIN \cite{Huang2017AdaIN}.

%

\paragraph{Stylizing the Output.}
Figure \ref{fig:spadestyleall}c illustrates the information flow from the encoder to the generator. We calculate the latent style code by sampling a set of images $I^{(1)}, I^{(2)},..., I^{(k)}$ ($k \in \mathbb{N}$) from a specified target person and embed them via an encoder network $\bv{s}^{(i)}=E_{style}(I^{(i)})$.
The style codes are aggregated by taking the element-wise maximum:
$\bv{s}_i = \max\limits_{1 \leq j \leq k} \bv{s}_i^{(j)}$.

This is inspired by the set-based face recognition literature \cite{Parkhi2015BMVC,Schroff2015CVPR}, where variations in appearance and head pose in the real world and consequent loss of information can be mitigated by merging information from multiple images of the same person.
Our style encoder uses spectral instance normalization in order to preserve style information \cite{Huang2017AdaIN,Li2017AAAI}.
The aggregated style code is used to calculate the parameters of AdaIN in the generator blocks. In this way, we can create a realistic eye image of a person with just a few unlabeled eye images and perform latent walks in style space.%


%

\paragraph{Content Preservation.} In order to produce images following the input segmentation mask, our generators repeatedly apply spatially adaptive normalization (SPADE) \cite{Park2019SPADE}. In comparison to batch or instance norm, where the adaptive parameters modify channels of feature maps as a whole, the SPADE layer learns a scale and offset for each element in the feature map based on the reference segmentation mask.

\paragraph{Generator Architecture.} 
Our generator model is an extension of GauGAN \cite{Park2019SPADE}, a model that applies spatially adaptive normalization (SPADE) to generate synthetic images given a semantic segmentation mask. We modify the SPADE block to inject both content and style. Our combined normalization, the SPADE+Style Block, takes three inputs: a semantic segmentation map $M$, a style vector $\bv{s}$ and the actual feature maps $X$. The input feature maps $X$ take two different paths and are merged again by addition before leaving the block.
One path is an AdaIN \cite{Huang2017AdaIN} layer where the parameters are computed from a style vector. The AdaIN layer applies a learned affine transform to adjust the dimensionality of the style vector to the correct number of channels. The other path is a SPADE block as described by \cite{Park2019SPADE}, but we apply spectral instance norm instead of synchronized batch norm. The output is divided by two, which we found to stabilize the training. In short, we compute $\text{SPADE+STYLE}(X) = (\text{SPADE}(X) + \text{AdaIN}(X)) /{2}$.


Following GauGAN, our generator starts from a small segmentation map and then doubles feature map resolution in a series of SPADE+Style ResBlocks. Each SPADE+Style ResBlock consists of two SPADE+Style Blocks and a residual connection. Figure \ref{fig:spadestyleall} shows the elements of a SPADE+Style (a) Block and (b) ResBlock.

\paragraph{Training.} We train \ganmethod on paired labeled samples from the training subset of OpenEDS, and sample style images from the top 200 images from the similarity ranking stage (Section~\ref{sec:methodnn}).
Further implementation information can be found in our supplementary materials.

\subsubsection{Objective Function}





\paragraph{Discriminator Losses.}
The adversarial loss $\mathcal{L_{GAN}}$ follows GauGAN \cite{Park2019SPADE}  
 %
 and thus takes as input the concatenation of the semantic segmentation mask and the generated image. We also apply an L1 consistency loss on the discriminator feature maps. Concretely, let $F_D^{(i)}(\cdot)$ extract the feature map of the discriminator layer $i$ and let $I$ and $\hat I$ be the real and generated image. Our feature map consistency loss is computed as the sum of the intermediate feature map consistency terms. Let $m$ be the number of feature maps in the discriminator. We compute the loss as
 \begin{equation} 
    \mathcal{L_{D_{F}}} = \Sigma_{i=2}^{m} ||F_D^{(i)}(\hat I) - F_D^{(i)}(I) ||_1
 \end{equation}

\paragraph{Pixel-Matching Loss.}
As we have paired training samples, we apply a simple L2 consistency loss on the generated vs. target image: $\mathcal{L}_{L2} ||\hat I - I||_2$.

\paragraph{Style Consistency Loss.}
We compute the style consistency loss by passing the generated image $\hat I$ through the style encoder $E_{style}$. 
Let $\bv{s}$ be the aggregated style vectors as described above and $ \bv{\hat s} = E_{style}(\hat I)$ be the style vector for the generated image $\hat I$. 
We compute the latent style code consistency loss as
$\mathcal{L}_{style} = ||\bv{s} - \bv{\hat s}||_2$. 
In addition, we compute a consistency loss on the Gram matrix of the feature maps of the encoder. 

Let $F_E^{(i)}(\cdot)$ extract the $i$-th feature map of the encoder $E_{style}$. We calculate the encoder Gram matrix consistency loss as
\begin{equation}
\mathcal{L}_{Gram} = \Sigma_{i = 2}^{m} ||Gr(F_E^{(i)}(\hat I)) - Gr(F_E^{(i)}(I)) ||_1
\end{equation}
where $m$ is the number of feature maps in the encoder and $Gr(\cdot)$ is the function to compute the Gram matrix \cite{Gatys2016CVPR}.

\paragraph{Full Training Objective.}
The full training objective for the generator $G$ (given a discriminator $D$) is written as:
\begin{equation}
\begin{split}
    \mathcal{L}_G
    =
    & \lambda_{GAN} \mathcal{L}_{GAN} 
    + \lambda_{D_{F}} \mathcal{L}_{D_{F}} 
    + \lambda_{L2} \mathcal{L}_{L2} \\
    &+ \lambda_{style}\mathcal{L}_{style} 
    + \lambda_{Gram}\mathcal{L}_{Gram}
\end{split}
\end{equation}

\noindent with $\lambda_{GAN}=\lambda_{D_{FM}}=10$, $\lambda_{L2}=15$, $\lambda_{style}=0.5$ and $\lambda_{Gram}=10^4$.

%% file: 06_results.tex
\section{Results}
The per-image objective function of the OpenEDS Synthetic Eye Generation Challenge\footnote{\scriptsize{\url{http://evalai.cloudcv.org/web/challenges/challenge-page/354/evaluation} }
}
is given as: $\frac{1}{HW} \sqrt{\Sigma_i^H \Sigma_j^W (\hat I_{ij} - I_{ij})^2}$. For this challenge, we developed multiple models, two of which we present in this paper. Section \ref{sec:resultsrefiner} describes the results that optimizes for the target metric and wins the challenge and Section \ref{sec:resultsgan} talks about an alternative solution to the problem.

\subsection{Refiner Network}
\label{sec:resultsrefiner}
\begin{figure}[tbp]
  \centering    
\includegraphics[width=\columnwidth]{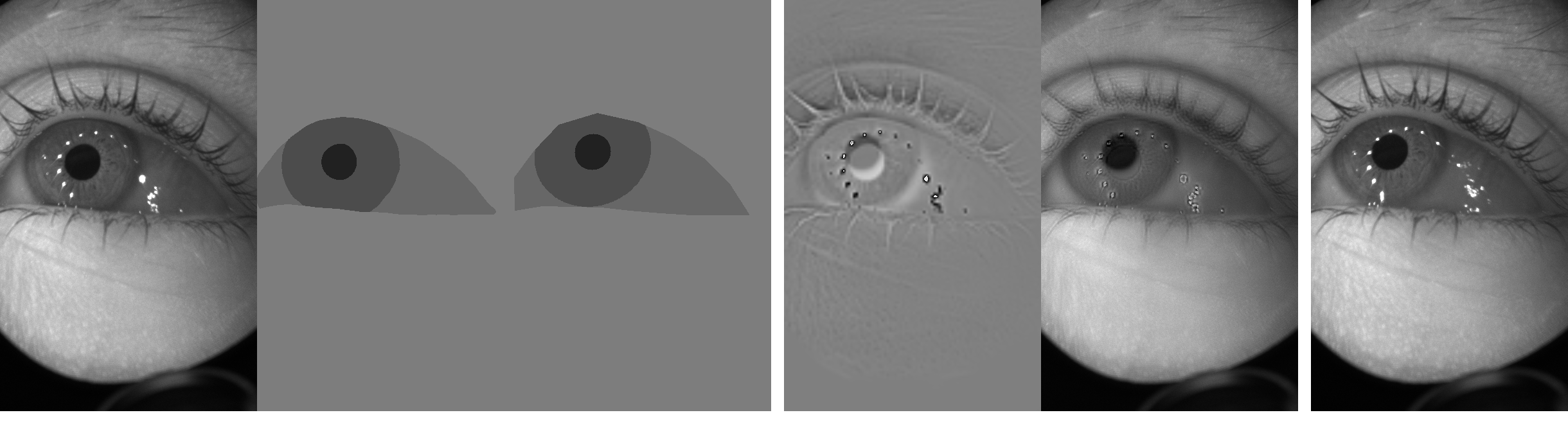}
    \vskip -4mm
    \caption{Qualitative outputs of the Refiner Network. The left three columns show the input reference image, pseudo-label and target label. The three columns on the right show the learned residual map, the final predicted image and the target image.
    We see that the modified regions are blurry due to optimizing for the L2 score. 
    }
    \label{fig:resultsrefiner}
    \vskip -4mm
\end{figure}

Our refinement model (cf. Section~\ref{sec:refiner}) achieved the lowest score of all teams at $25.23$. The scores of the second (PAU) and third teams (tomcarrot) were $27.69$ and $33.79$, respectively. The baseline provided by the challenge organizers was $59.25$.
Although the refiner network could win the challenge, we found that its generated images contain blurry regions and ghosting effects (see Figure~\ref{fig:resultsrefiner}). We believe that the reason for these effects lie in the L2 optimization objective, which encourages ``washed out" regions.


\begin{figure}[tbp]
  \centering    
\includegraphics[width=\columnwidth]{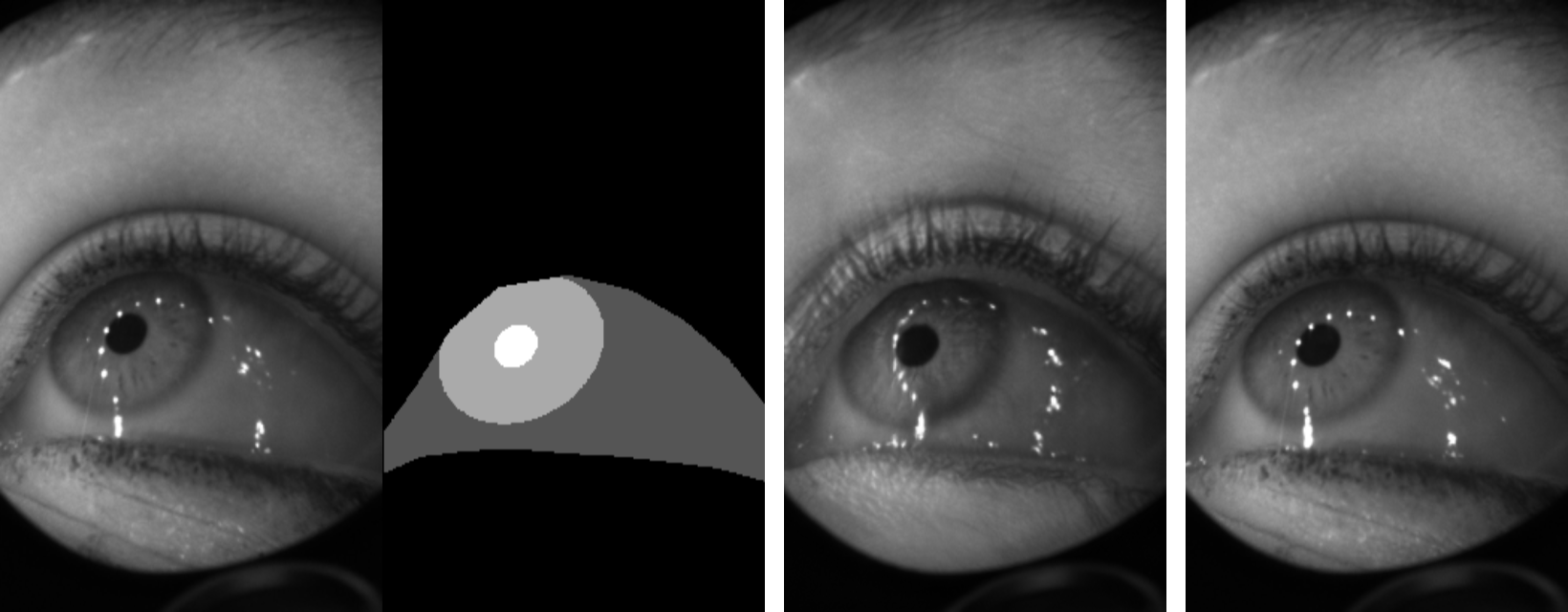}
    \vskip -3mm
    \caption{Qualitative outputs of \ganmethod. From left to right are (1) one of the style image inputs, (2) target segmentation mask input, (3) generated image, and (4) target real image taken from the validation set.
    The generated image closely follows the segmentation mask and input style.
    }
    \label{fig:resultsgan}
    \vskip -4mm
\end{figure}

\subsection{\ganmethod}
\label{sec:resultsgan}
The original GauGAN architecture starts from just a single downsampled style image, whose style is supposed to be preserved. When directly applied to the OpenEDS dataset, we found that all output images followed the same style, i.e., a complete lack of style preservation. We introduced our novel SPADE+Style blocks to incorporate style information via AdaIN, specifically allowing for the merging of style information from several images via a style encoder and max-reduction. We found that this improves model performance in terms of both L2 score and visual quality. However, without any additional encouragement, the network did not learn a consistent latent style space that allowed for interpolation in the style latent space. For this reason, we added a style consistency loss on both the latent code and the Gram matrix of the encoder feature maps. In this scenario, style was applied considerably better and we could perform latent space walks as shown in Figure \ref{fig:interpolation}  

The final \ganmethod approach does not achieve the lowest score of all approaches, but produces realistic images of high-perceptual quality. Figure \ref{fig:resultsgan} illustrates some example in- and outputs. 
It can be seen that visual fidelity is vastly improved compared to the Refiner Network.
Additionally, Figure \ref{fig:interpolation} shows a style interpolation between two people, demonstrating that a good understanding of style has been met.
We also believe that this is more in line with the expected final usage of such a method in generating vast amounts of training data in a controlled manner.

%% file: 07_conclusion.tex
\section{Conclusion}
In this paper, we described both the winning approach to the OpenEDS Synthetic Eye Generation Challenge and a principled approach to generating person-specific eyes given a semantic segmentation map. Our method, \ganmethod, is inspired by \cite{karras2018stylegan} and suggests a modification to spatially adaptive normalization as introduced by \cite{Park2019SPADE} that leads to a a more consistent application of style.

In future work, one should explore different ways to combine the content and style information in the \ganmethod generator, or combine style information from multiple style images. Furthermore, to truly take advantage of \ganmethod, an interpretable latent space of eye shapes should be learned such that plausible and high fidelity IR images of eyes can be created from entirely unseen people.

\paragraph{Acknowledgements.}
This work was supported in part by the ERC Grant OPTINT (StG-2016-717054).

%% file: ms.bbl
\begin{thebibliography}{10}\itemsep=-1pt

\bibitem{Chen2016DeepLab:CRFs}
Liang-Chieh Chen, George Papandreou, Iasonas Kokkinos, Kevin Murphy, and Alan~L
  Yuille.
\newblock Deeplab: Semantic image segmentation with deep convolutional nets,
  atrous convolution, and fully connected crfs.
\newblock {\em TPAMI}, 40(4):834--848, 2017.

\bibitem{chen2017rethinking}
Liang-Chieh Chen, George Papandreou, Florian Schroff, and Hartwig Adam.
\newblock Rethinking atrous convolution for semantic image segmentation.
\newblock {\em arXiv preprint arXiv:1706.05587}, 2017.

\bibitem{Ganin2016ECCV}
Yaroslav Ganin, Daniil Kononenko, Diana Sungatullina, and Victor Lempitsky.
\newblock Deepwarp: Photorealistic image resynthesis for gaze manipulation.
\newblock In {\em ECCV}, pages 311--326, 2016.

\bibitem{Garbin2019arXiv}
Stephan~J. Garbin, Yiru Shen, Immo Schuetz, Robert Cavin, Gregory Hughes, and
  Sachin~S. Talathi.
\newblock Openeds: Open eye dataset.
\newblock {\em CoRR}, abs/1905.03702, 2019.

\bibitem{Gatys2016CVPR}
Leon~A Gatys, Alexander~S Ecker, and Matthias Bethge.
\newblock Image style transfer using convolutional neural networks.
\newblock In {\em CVPR}, pages 2414--2423, 2016.

\bibitem{He2019ICCV}
Zhe He, Adrian Spurr, Xucong Zhang, and Otmar Hilliges.
\newblock Photo-realistic monocular gaze redirection using generative
  adversarial networks.
\newblock In {\em ICCV}, 2019.

\bibitem{Huang2017AdaIN}
Xun Huang and Serge Belongie.
\newblock Arbitrary style transfer in real-time with adaptive instance
  normalization.
\newblock In {\em ICCV}, 2017.

\bibitem{Huang2018ECCV}
Xun Huang, Ming-Yu Liu, Serge Belongie, and Jan Kautz.
\newblock Multimodal unsupervised image-to-image translation.
\newblock In {\em ECCV}, September 2018.

\bibitem{isola2017pix2pix}
Phillip Isola, Jun-Yan Zhu, Tinghui Zhou, and Alexei~A. Efros.
\newblock Image-to-image translation with conditional adversarial networks.
\newblock In {\em CVPR}, July 2017.

\bibitem{karras2018stylegan}
Tero Karras, Samuli Laine, and Timo Aila.
\newblock A style-based generator architecture for generative adversarial
  networks.
\newblock {\em CoRR}, abs/1812.04948, 2018.

\bibitem{Kim2019CHI}
Joohwan Kim, Michael Stengel, Alexander Majercik, Shalini De~Mello, David Dunn,
  Samuli Laine, Morgan McGuire, and David Luebke.
\newblock Nvgaze: An anatomically-informed dataset for low-latency, near-eye
  gaze estimation.
\newblock In {\em CHI}, 2019.

\bibitem{Lee2018ICLR}
Kangwook Lee, Hoon Kim, and Changho Suh.
\newblock Simulated+unsupervised learning with adaptive data generation and
  bidirectional mappings.
\newblock In {\em ICLR}, 2018.

\bibitem{Li2017AAAI}
Yanghao Li, Naiyan Wang, Jiaying Liu, and Xiaodi Hou.
\newblock Demystifying neural style transfer.
\newblock In {\em IJCAI}, pages 2230--2236, 2017.

\bibitem{Park2019ICCV}
Seonwook Park, Shalini~De Mello, Pavlo Molchanov, Umar Iqbal, Otmar Hilliges,
  and Jan Kautz.
\newblock Few-shot adaptive gaze estimation.
\newblock In {\em ICCV}, 2019.

\bibitem{Park2018ETRA}
Seonwook Park, Xucong Zhang, Andreas Bulling, and Otmar Hilliges.
\newblock Learning to find eye region landmarks for remote gaze estimation in
  unconstrained settings.
\newblock In {\em ACM Symposium on Eye Tracking Research and Applications
  (ETRA)}, 2018.

\bibitem{Park2019SPADE}
Taesung Park, Ming-Yu Liu, Ting-Chun Wang, and Jun-Yan Zhu.
\newblock Semantic image synthesis with spatially-adaptive normalization.
\newblock In {\em CVPR}, 2019.

\bibitem{Parkhi2015BMVC}
Omkar~M Parkhi, Andrea Vedaldi, and Andrew Zisserman.
\newblock Deep face recognition.
\newblock In {\em BMVC}, 2015.

\bibitem{Schroff2015CVPR}
Florian Schroff, Dmitry Kalenichenko, and James Philbin.
\newblock Facenet: A unified embedding for face recognition and clustering.
\newblock In {\em CVPR}, 2015.

\bibitem{Sela2017arXiv}
Matan Sela, Pingmei Xu, Junfeng He, Vidhya Navalpakkam, and Dmitry Lagun.
\newblock Gazegan - unpaired adversarial image generation for gaze estimation.
\newblock {\em CoRR}, abs/1711.09767, 2017.

\bibitem{Shrivastava2017CVPR}
Ashish Shrivastava, Tomas Pfister, Oncel Tuzel, Joshua Susskind, Wenda Wang,
  and Russell Webb.
\newblock Learning from simulated and unsupervised images through adversarial
  training.
\newblock In {\em CVPR}, July 2017.

\bibitem{wang2018pix2pixhd}
Ting-Chun Wang, Ming-Yu Liu, Jun-Yan Zhu, Andrew Tao, Jan Kautz, and Bryan
  Catanzaro.
\newblock High-resolution image synthesis and semantic manipulation with
  conditional gans.
\newblock In {\em CVPR}, June 2018.

\bibitem{wood2018gazedirector}
Erroll Wood, Tadas Baltru{\v{s}}aitis, Louis-Philippe Morency, Peter Robinson,
  and Andreas Bulling.
\newblock Gazedirector: Fully articulated eye gaze redirection in video.
\newblock In {\em Computer Graphics Forum}, volume~37, pages 217--225. Wiley
  Online Library, 2018.

\bibitem{Yu2019CVPR}
Yu Yu, Gang Liu, and Jean-Marc Odobez.
\newblock Improving few-shot user-specific gaze adaptation via gaze redirection
  synthesis.
\newblock In {\em CVPR}, June 2019.

\end{thebibliography}


\begin{thebibliography}{1}\itemsep=-1pt

\bibitem{Chen2016DeepLab:CRFs}
Liang-Chieh Chen, George Papandreou, Iasonas Kokkinos, Kevin Murphy, and Alan~L
  Yuille.
\newblock Deeplab: Semantic image segmentation with deep convolutional nets,
  atrous convolution, and fully connected crfs.
\newblock {\em TPAMI}, 40(4):834--848, 2017.

\bibitem{Kingma2014Adam:Optimization}
Diederik~P. Kingma and Jimmy Ba.
\newblock {Adam: A Method for Stochastic Optimization}.
\newblock 12 2014.

\bibitem{Park2019SPADE}
Taesung Park, Ming-Yu Liu, Ting-Chun Wang, and Jun-Yan Zhu.
\newblock Semantic image synthesis with spatially-adaptive normalization.
\newblock In {\em CVPR}, 2019.

\end{thebibliography}
